\documentclass[conference]{IEEEtran}
\IEEEoverridecommandlockouts
\usepackage{cite}
\usepackage{amsmath,amssymb,amsfonts}
\usepackage{algorithmic}
\usepackage{graphicx}
\usepackage{textcomp}
\usepackage{xcolor}
\usepackage{multirow}
\def\BibTeX{{\rm B\kern-.05em{\sc i\kern-.025em b}\kern-.08em
    T\kern-.1667em\lower.7ex\hbox{E}\kern-.125emX}}

\usepackage[utf8]{inputenc} 
\usepackage[T1]{fontenc}    
\usepackage{hyperref}       
\usepackage{url}            
\usepackage{booktabs}       
\usepackage{amsfonts}       
\usepackage{nicefrac}       
\usepackage{microtype}      

\usepackage[labelfont=bf]{caption}
\usepackage{enumitem}
\usepackage{mathtools}
\usepackage{subfig}

\usepackage[ruled,vlined]{algorithm2e}

\usepackage{lscape}
\usepackage{pifont}
\usepackage{multirow}

\newcommand{\proj}{FedDP\xspace} 

\begin{document}

\title{Heterogeneous Federated Learning using Dynamic Model Pruning and Adaptive Gradient\\
}

\author{

\IEEEauthorblockN{
Sixing Yu \IEEEauthorrefmark{1}\textsuperscript{\textsection},
Phuong Nguyen\IEEEauthorrefmark{1}\textsuperscript{\textsection}, 
Ali Anwar\IEEEauthorrefmark{2} and
Ali Jannesari\IEEEauthorrefmark{1}
}

\IEEEauthorblockA{
\IEEEauthorrefmark{1} Department of Computer Science,
Iowa State University
}
\IEEEauthorblockA{
\IEEEauthorrefmark{2} Department of Computer Science \& Engineering,
University of Minnesota
}
\IEEEauthorblockA{
yusx@iastate.edu,
dphuong@iastate.edu,
aanwar@umn.edu,
jannesar@iastate.edu
}
}

\maketitle

\begingroup\renewcommand\thefootnote{\textsection}
\footnotetext{Equal contribution}
\endgroup


\begin{abstract}
Federated Learning~(FL) has emerged as a new paradigm for training machine learning models distributively without sacrificing data security and privacy. Learning models on edge devices such as mobile phones is one of the most common use cases for FL. 
However, Non-identical independent distributed~(non-IID) data in edge devices easily leads to training failures. Especially, over-parameterized machine learning models can easily be over-fitted on such data, hence, resulting in inefficient federated learning and poor model performance. 
To overcome the over-fitting issue, we proposed an adaptive dynamic pruning approach for FL, which can dynamically slim the model by dropping out unimportant parameters, hence, preventing over-fittings. Since the machine learning model's parameters react differently for different training samples, adaptive dynamic pruning will evaluate the salience of the model's parameter according to the input training sample, and only retain the salient parameter's gradients when doing back-propagation. 
We performed comprehensive experiments to evaluate our approach. The results show that our approach by removing the redundant parameters in neural networks can significantly reduce the over-fitting issue and greatly improves the training efficiency. In particular, when training the ResNet-32 on CIFAR-10, our approach reduces the communication cost by 57\%. We further demonstrate the inference acceleration capability of the proposed algorithm. Our approach reduce up to 50\% FLOPs inference of DNNs on edge devices while maintaining the model's quality. 

\end{abstract}

\section{Introduction}

With the increasing concern about user-privacy data, training AI models on edge devices without private data access has become a challenge. 
Federated learning~(FL) is developed as a decentralized machine learning approach to not only cope with the privacy issue but also to efficiently optimize AI models on imbalanced distributed data~\cite{mcmahan2017fedavg}.
Intuitively, FL aims to optimize deployed AI models in a distributed manner and aggregate decentralized training results on a cloud/central server without touching private data.
Recently, numerous algorithms for FL and its variants have been proposed, such as SCAFFOLD~\cite{karimireddy2020scaffold} and FedNova~\cite{wang2020fednova}, which significantly stabilize the optimizing process of the original FL algorithm, i.e., FedAvg~\cite{mcmahan2017fedavg}.
However, regardless of these novel ideas, the issue of imbalanced and non-identical independent distributed data~(Non-IID) still leads to the training failure of FL.
Since the model is trained in a decentralized setting without access and understanding of the whole training data in the FL setting, and the data on the local device is inherently non-IID, the model performance is not as high as that trained in the centralized setting.

We observe that a critical factor explaining the poor converging accuracy of FL that has been overlooked by researchers so far is the over-fitting issue, which significantly reduces the convergence speed and increases unexpected training fluctuations.
Since modern AI models deployed on the decentralized system are typically over-parameterized~(e.g., deep neural network), the tiny and imbalanced local data at the edges can easily be over-fitted.
An effective way to address over-fitting is to increase the amount of data. Nevertheless, in the FL scenario, the local data is created by the user and has privacy restrictions, and most importantly, edge devices have limited computational and storage resources. Therefore, it is not practical to augment data on edge clients.
Another solution for over-fittings is to slim the model size. However, the non-IID and imbalanced data in edge clients raise challenges. For instance, some clients may have plenty of training samples to train the model well, while other clients may have limited training samples to get over-fitting. Hence, shrinking the model size mechanically would lead to unfairness and biases among edge clients.

In this paper, we come up with an approach to slim the model dynamically.
In a neural network, neurons respond differently depending on different inputs. For a given batch of data, only a subset of salient parameters contributes to the final prediction. For any input sample, we dynamically select the corresponding salient parameter and retain the gradient to update.
We propose adaptive \underline{D}ynamic \underline{P}runing for efficient \underline{Fed}erated learning~(\proj). 
Specifically, we introduce a dynamic pruning component on each edge device. The component selectively predicts the salient parameters for every given training sample, which masks the unimportant and redundant parameters of the given training sample and only calculates the gradients for the remaining important parameters. Moreover, to further prevent the gradient drift among heterogeneous edge devices, we leverage global gradient control algorithm that corrects the heterogeneity and guides the gradient toward a generic global direction among all edge devices. 

In essence, this paper makes the following contributions:
\begin{itemize}
    \item Mitigating the over-fitting issue on edge clients through a novel adaptive pruning technique.
    \item An adaptive network pruning method for non-IID edge devices with no additional communication cost.
    \item Dataset-aware pruning policy for each edge device on federated learning network.
    \item Reduction up to $2\times$ on communication cost and 50\% FLOPs on local inference.  
\end{itemize}

\begin{figure*}
\begin{center}
  \includegraphics[width=.7\linewidth]{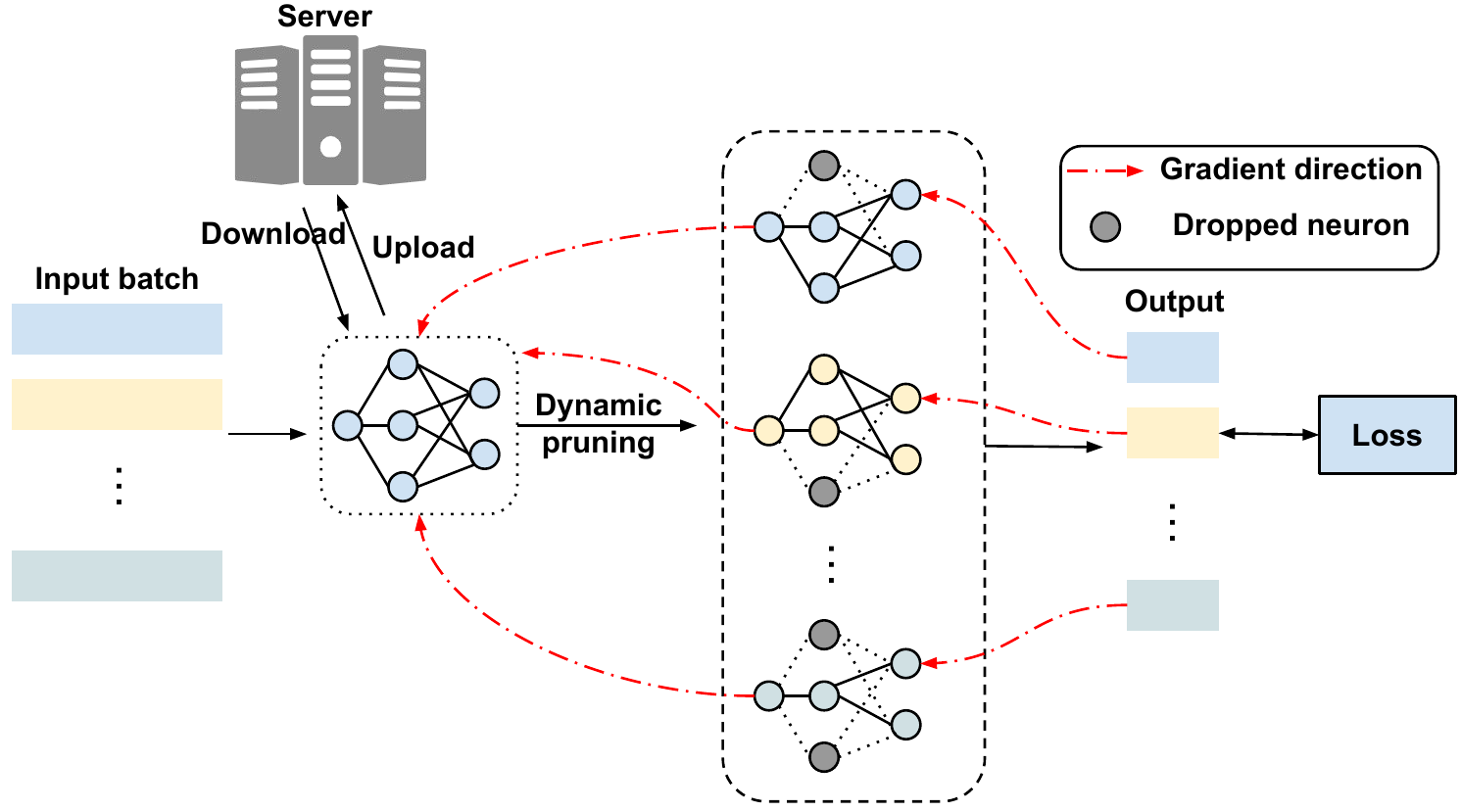}
\end{center}
   \caption{Adaptive gradient control local updates. }

\label{fig:dp}
\end{figure*}
\section{Motivation}
Nowadays, edge applications are ubiquitous with the popularity and enhanced performance of mobile and embedded devices. Improved ML models and tremendous data created on edge mean a remarkable amount of future ML workload will come from training and optimization on edge devices. 
However, great challenges are coming from edge devices due to strict imbalanced data distribution, and privacy constraints, which invalidate centralized training ML models and cause training failures.
Therefore, the prevailing edge computing trend alongside FL requirements and edge constraints motivates \proj to address challenges in HPC.

Firstly, decentralized training on edge devices needs significantly more training rounds than centralized training. Besides, Optimizing ML model on edge devices requires frequent sharing of model weights between edge devices and the central server. These incur hefty communication costs between edge devices and the server. 


Secondly, the optimization process in over-parameterized ML~(e.g. deep neural networks) models require mass computing, memory, and storage resources, which are hard to accomplish on resource-constrained Internet-of-Things~(IoT) and edge devices~\cite{imteaj2022iot,nguyen2021iot}. Hence, reasonably pruning redundant weights and dropping out unnecessary gradients can be a viable solution to efficiently optimize ML models.


Thirdly, with the increasing concern of privacy issues in the digital world, how to reasonably utilize these private data on such privacy sensitive-manner proves to be challenging.
Therefore, efficiently optimizing ML models without directly accessing user-private data is important in FL and HPC settings.

Lastly, tech giants~(such as Google, Apple, and NVIDIA) are actively using FL for their applications (e.g., Google Keyboard~\cite{yang2018keyboard,chen2019keyboard}, Apple Siri~\cite{apple2019siri,paulik2021siri}, NVIDIA medical imaging~\cite{li2019nvidia}). The huge number of clients produces an incredible amount of data every second. The imbalanced and heterogeneous data created on edge amplify many issues for decentralized model optimization. As a result, designing dynamic and adaptive algorithms to process the data will be important to improve the performance of future edge computing.


\section{Related work}

\subsection{Federated learning}
Federated learning is a decentralized machine learning approach by communication with the model parameters by leaving the training data on each client's devices in order to keep the clients' privacy~\cite{lim2020federated}. 
General FL involves two entities, the server and the clients (edges), and consists of three phases such as downstream, upstream, and edge computation~\cite{mcmahan2017fedavg}. 
The downstream means transmitting the initialized model from the server to clients and then training the model on each local data in the edge computation phase. After that, the clients have the upstream phase uploading the weights (parameters) trained on each client to the server and aggregating the local models.
The key idea is to train the model on each local device, which results in the computation and storage limitation on the edges' device. The compression scheme is one of the directions in order to improve communication efficiency. 
Kone{\v{c}}n{\`y} et al.~\cite{konecny2016federated_FLquantization} proposed structure updates and sketched updates in the client upstream phase by enforcing the size of update parameters and sub-sampling the update using quantization and random rotations, respectively. 
Caldas et al.~\cite{caldas2018expanding_FLquantization} suggested adopting the lossy compression on the downstream phase from server to clients and dropout on the upstream phase and local round in order to decrease the communication cost.

Reducing the computation on clients' devices is also related to achieving the efficient strategy of FL~\cite{Li2020Federated}. Moreover, theoretical schemes, following Independent and Identical Distributed~$($IID$)$ data, could not be applied in a practical environment~\cite{lim2020federated}. Therefore, the non-IID issue is also one of the challenges in federated learning because data on each device is following different distributions.
McMahan et al.~\cite{mcmahan2017fedavg} proposed Federated Averaging~(FedAvg) to resolve the non-IID issue by iterative model averaging with local stochastic gradient descent~(SGD) on each local client's data. Moreover, they demonstrated the robustness of FedAvg to train convolutional neural networks~(CNNs) on benchmark image classification datasets (e.g., MNIST~\cite{lecun98mnist}
and CIFAR-10~\cite{krizhevsky2009cifar}). Other state-of-the-art~(SoTA) algorithms includes FedProx~\cite{li2020fedprox}, FedNova~\cite{wang2020fednova} and SCAFFOLD~\cite{karimireddy2020scaffold}. These are all variants from FedAvg. For instance, FedProx restricts local update sizes while FedNova introduces weight modification to improve the aggregation phase and SCAFFOLD corrects update direction by maintaining a drift in local training.

\subsection{Non-Independent and Identical Distributed Data (non-IID)}
Non-IID data issue or heterogeneity is one of the main challenges in FL~\cite{problemsfl,nishio2019clientselection,zhao2018federated,hsieh2020quagmire,li2020convergence_noniid,lim2020federated,li2020fedprox,reddi2021fedopt,karimireddy2020scaffold,wang2020fednova,zhang2021adaptive_noniid}. 
There are two types of heterogeneity: (a) statistical heterogeneity, and (b) system heterogeneity~\cite{li2020fedprox}. Heterogeneity affects various aspects of the machine learning pipeline in FL. It can lead to slower convergence, reduced stability of convergence procedure, and divergence~\cite{karimireddy2020scaffold,li2020convergence_noniid,yang2020imbalance,li2020fedprox}. Another problem is we might perform very well on average but in general, there are no accuracy guarantees for individual devices and the performance may therefore vary wildly across the network~\cite{li2020fair, hashimoto2018fairness, mohri2019agnostic, li2021_term}. Finally, modeling is an overarching issue~\cite{smith2017_mtl}. Traditionally, the goal is to learn one model across the entire network, when data is very heterogeneous, it can be important to consider models that go beyond that and can personalize to data that's being generated on these devices~\cite{smith2017_mtl}.
To improve FL with model compression from non-IID distribution, Sattler et al.~\cite{sattler2019robust} propose the communication efficient FL with non-IID data distribution by compressing the model on both phases, upstream and downstream, using k-sparsification. 
Xu et al.~\cite{xu2021accelerating} suggested accelerating the FL model with three stages: pruning, quantization, and selective updating. However, the pruning is to adopt structured pruning as a static pruning model. This limits the flexibility of the non-IID issue because the structured pruning reduces the model size without considering the current input. This paper focuses on statistical heterogeneity.\\

\subsection{Model compression}
Model compression such as network pruning~\cite{li2020eagleeye, ye2020goodsubnet,yu2020agmc,yu2021gnnrl}, knowledge distillation~\cite{chen2017learning}, and network quantization~\cite{polino2018model} focus on efficient deployment of deep neural networks~(DNNs) on edge devices. Within the scope of this paper, we mainly discuss network pruning. 
Network pruning can be categorised to static pruning~\cite{chin2020towards, guo2020dmcp, li2020eagleeye} and dynamic pruning~\cite{li2021dynamic,gao2018dynamic}.
Static pruning evaluates model parameters' importance and removes those with a lower rank permanently. Static pruning can reduce the model size but will permanently lose pruned parameter's information. Otherwise, the static pruning policy is highly input-dependent, a universal pruning policy for non-IID edge devices is not practical.
Dynamic pruning, instead of simply reducing model size at the cost of accuracy with pruning, accelerate convolution by selectively computing only a subset of channels predicted to be important at run-time while considering the sparse input from the preceding convolution layer. In addition to saving computational resources, a dynamic model preserves all neurons of the full model, which minimizes the impact on task accuracy.
However, only dynamic pruning fails to shrink the model size and will not improve communication efficiency.

The limited computing resources of edge devices on federated learning and the frequent communication between edge devices and server makes FL more challenging. Resolving these issues is vital for a successful FL and further applications. Only a limited number of research try to address the above issues by performing model compression techniques on edge devices (e.g., pruning on edge devices PruneFL~\cite{jiang2020pruneFL} or model/gradient quantization compression techniques~\cite{xu2019elfish_FLquantization,han2020adaptive_FLquantization,caldas2018expanding_FLquantization,konecny2016federated_FLquantization}). However, there is no available solution to perform both the communication overhead and the limited computing resource issue simultaneously. 


\section{Approach}

\begin{figure}
\begin{center}
  \includegraphics[width=\linewidth]{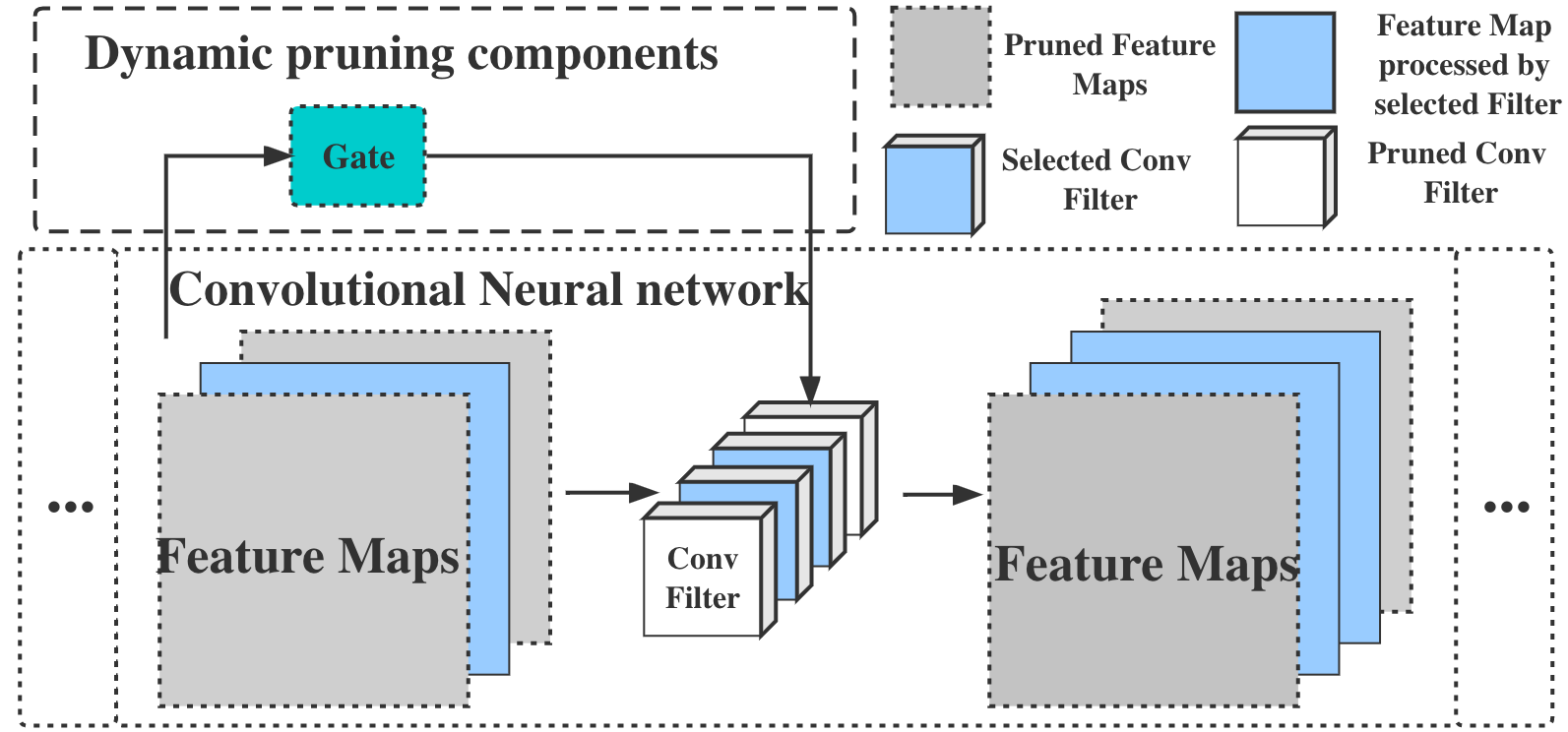}
\end{center}
   \caption{Dynamic pruning components for CNN. }

\label{fig:dp_cnn}
\end{figure}
Figure~\ref{fig:dp} illustrates the high-level overview of \proj. The edge clients first download the model from the server and then optimize it on the local Non-IID dataset. For each input training sample, \proj dynamically drops off unimportant parameters when calculating the gradient according to the given sample. 
The dynamic pruning component is used to adaptively prune redundant and unimportant weights, which have a low response to the input sample.
Hence, when performing back-propagate, those dropped parameters and corresponding feature maps would not retain gradients. 
Dynamically pruning/dropping out redundant parameters makes the model robust and easy to converge, which can significantly relieve the over-fitting issue and accelerate the model inference on edge devices. 

\subsection{Adaptive dynamically pruning}
To prevent the model suffer from over-fittings on non-IID local datasets,
we apply adaptive dynamic pruning to selectively drop off unimportant parameters based on the parameter's reactions to the input training sample.

To intuitively explain our approach, as depicted in Figure~\ref{fig:dp_cnn}, we take the CNNs as an example.
To dynamically prune a neural network, we introduced a dynamic pruning gate for each hidden layer. The gate will take the feature maps~(previous layer's output) as input, predict which channels~(Conv Filter) are important in the current layer, and drop out unimportant parameters. 

Formally, we formulate a $l-$layer CNN as equation~\ref{eq:cnn}.
\begin{equation}
\label{eq:cnn}
    CNN(x) = f_{l}(f_{l-1}(...f_{1}(x_0)...))
\end{equation}
, where the $x_i$ is the feature map in $i^{th}$ layer, $CNN(x)$ is the forward propagate function of the neural network, and the $f_{k}$ is the $k^{th}$ hidden layer, which projects the feature maps from $(k-1)^{th}$ layer to $k^{th}$ layer. In CNN model, a commonly used hidden layer $f_{k}$ can be represented as equation~\ref{eq:h_layer}.
\begin{equation}
    \label{eq:h_layer}
    f_{k}(x_{k-1})=ReLu(norm(Conv_{k}(x_{k-1},\theta_{k})))
\end{equation}
, where the $\theta_{k} \in \mathbb{R}^{C_{out}\times C_{in}\times K\times K}$ is the learnable parameter, $C_{out}$ and $C_{in}$ is the output and input channel, K is the kernel size, $Conv_{k}$ is the convolutional operation, the norm is the normalization, and ReLu is the activation function.

In CNNs, each hidden layer can be viewed as a feature extractor, and it would respond differently according to input samples. We introduced dynamic pruning gates among each hidden layer to capture hidden layers' reactions with different inputs. The dynamic pruning gate will adaptively drop off low response parameters. 
For example, in the $k$ hidden layer, the gate convolutional layer is defined as equation~\ref{eq:slicing} and~\ref{eq:gateconv}.
\begin{equation} \label{eq:slicing}
    \hat{\theta}_{k} = \theta_{k}[Gate_k(x_{k-1}),Gate_{k-1}(x_{k-2}),:,:]
\end{equation}
\begin{equation}
    \hat{f_{k}}(x_{k-1}) = ReLu(norm(Conv_{k}(x_{k-1},\hat{\theta}_{k})))  
    \label{eq:gateconv}
\end{equation}
The $Gate_k(x_{k-1})$ predicts the important channels and produces the channel index tensor with shape $\mathbb{R}^{\hat{C}_{out} \times 1}$, where the $\hat{C}_{out}$ is the number of channels predicted to be important for k-th convolution layers. In the gate convolutional layer, we have the selected parameter $\hat{\theta}_{k}$ with size $\mathbb{R}^{\hat{C}_{out}\times \hat{C}_{in}\times K\times K}$.


In the back-propagation, only important parameters would be updated, as unimportant parameters have been dropped out and no gradient retained for them. Equation~\ref{eq:loss1} shows the local updates objective function.
\begin{equation}
\label{eq:loss1}
    \min_{w} \mathcal{L}(w) = \frac{1}{n_i}\sum l(w_i,x_i,y_i)
\end{equation}
, $l$ refers to the loss when fitting the label $y_i$ for data $x_i$, and $n_i$ is the constant coefficient, $w_i$ is the salient parameter for the input $x_i$.

\subsection{Dynamic pruning gate}
The dynamic pruning gate takes the current layer's feature map as input and predicts the important channels~(convolutional filters) to be selected.
Essentially, the dynamic pruning gate sub-samples the spatial dimensions of the feature map to scalars and a linear layer to make the prediction. The $Gate_k(x_{k-1})$ can be formulate as equation~\ref{eq:gate1} and ~\ref{eq:gate2}.
\begin{equation}
    \label{eq:gate1}
    s_k = \operatorname{subsample}(x_k) \in \mathbb{R}^{\hat{C}_{in}\times 1\times 1} 
\end{equation}
\begin{equation}\label{eq:gate2}
    g_k = {\operatorname{K-argmax}}(\operatorname{Sigmoid}(\operatorname{linear}(s_k))) \in \mathbb{R}^{\hat{C}_{out}\times 1\times 1} 
\end{equation}

\begin{algorithm}
    \caption{\proj}
    \label{alg:aggregate}
\begin{algorithmic}

\STATE {\bfseries Input:} $K$ clients indexed by $k$, number of communication rounds $T$, number of local epochs $E$, learning rate $\eta$, total number of samples $n$, number of sample in $k^{th}$ client $n_k$.
\STATE {\bfseries Output:} final model $w^T$.

\STATE

\STATE {\bfseries Server executes:} 
\STATE initialize $w^0,c_g$ 
\FOR{each round $t =0, 1, \ldots, T$}
    
    \STATE $K \leftarrow$ random set of clients
    \FOR{each client $k \in K$ {\bfseries in parallel}}
        \STATE send $w^{t}$ to client $k$
        \STATE $\hat{w}^{t+1}_k,{}{\hat{c}}_k\leftarrow$ ClientUpdate$(k,w^t,{}{c_g})$
    \ENDFOR
    
    \STATE $w^{t+1} \leftarrow  w^t + \sum_{k=1}^{K}\frac{n_k}{n} \hat{w}^t_k $  
        
    {$c_g \leftarrow  c_g + \frac{1}{|N|}\sum_{k \in K} \Delta \hat{c}_k $}

\ENDFOR




\STATE
\STATE {\bfseries ClientUpdate($k, w^t, {c_g}$):}
\STATE $\mathcal{B} \leftarrow$ split local dataset into batches 
\STATE $w_k^t, {c_g} \leftarrow$ download global model parameter $w^t$, and gradient control variates ${c_g}$
\STATE $c_l\leftarrow$ initialize local control $c_l$


\FOR{epoch$=1,2,\ldots, E$}
    \FOR{batch $b\in \mathcal{B}$}
        
        \STATE $w^t_k \leftarrow w^t_k - \eta \triangledown\mathcal{L}((w^t_k,w_g;b) {}{ - c_l + c_g}) $  \# update neural network
        \STATE $w_g \leftarrow w_g - \eta \triangledown\mathcal{L}(w^t_k,w_g;b)$  \# update pruning gate 
    \ENDFOR
 
\ENDFOR





{$c_l^* \leftarrow  c_l - c_g + \frac{1}{E\eta}(w^t - w_k^t)$}

$\hat{c}_k \leftarrow (c_l^* -c_l)$

{$c_l \leftarrow c_l^*$}

\STATE \textbf{communicate} $w^t_k,  {\hat{c}_k}$ to server

\end{algorithmic}
\end{algorithm}

\label{sec:dpg}

In the implementation, we use the average pool to subsample the feature map, and the $K$-argmax operation will return the top-$K$ most significant value's index. The $K$ is a hyper-parameter, which defines the dynamic channel pruning ratio for hidden layers.
Similar to the~\cite{gao2018dynamic}, we train the gate together with the neural network, and regularize all layers with the Lasso $\mathcal{L}_g = \lambda \sum_{k=1}^l \mathbb{E}[||\operatorname{linear(s_k)}||_1] $ in the total loss, where $s_k$ is the subsample vector of $x_k$.

\begin{figure*}
\begin{center}
 \vspace{-6 px}

\centerline{\includegraphics[width=\linewidth]{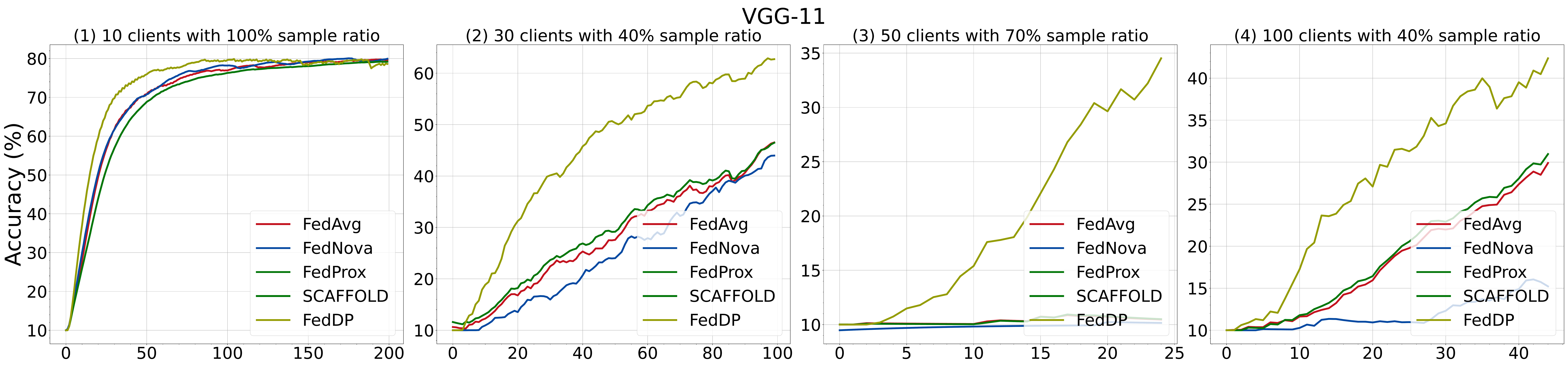}}

\caption{Comparison of FedDP with SoTA: the top-1 average test accuracy vs. communication rounds.}
 \vskip -0.3in
\vspace{-6 px}
\label{fig:scaffold_dp}
\end{center}
 \end{figure*}

\subsection{Global gradient control variates}
In \proj, we use the SCAFFOLD~\cite{karimireddy2020scaffold} as the backend FL aggregating algorithm.
Straightforwardly, we maintain a gradient control variate to stabilize the federated learning process. The variate in the cloud server tracks the global gradient direction among heterogeneous edge clients. The edge clients will use the gradient control variates to correct their gradient drift when performing local updates.

Algorithm~\ref{alg:aggregate} shows \proj with gradient controlled FL.
In each update round, the client downloads the global model's parameter $w^t$ and the global optimization direction $c_g$ from the server and performs local updates. 
In local clients, we maintain a local optimization control variate $c_l$,
and when training the local model $w_k^t$, $(c_g - c_l)$ is applied to correct the gradient drift. After the local training is finished, the local control variate $c_l$ is updated by estimating the gradient drift between local and global models.

\subsection{Adaptive gradient control federated learning}
Inspired by stochastic controlled averaging federated learning~\cite{karimireddy2020scaffold}, we equipped dynamic pruning with gradient-controlled federated learning to enable a much more stable and robust federated learning process.
Due to client heterogeneity, local gradient update directions will move towards local optima and may diverge across all clients. To correct overall gradient divergence by estimating gradient update directions, we maintain control variates both on clients and the cloud aggregator. 
This section shows how we equip adaptive dynamic pruning to gradient control federated learning.
Algorithm~\ref{alg:aggregate} shows the local updates and server aggregates process.


In the server execution, the procedure is exactly the same as the SCAFFOLD~\cite{karimireddy2020scaffold}, where the server aggregates the parameter from the selected clients. When doing the local update, the client first downloads the neural network's parameter $w^t$ from the cloud server at the $t$ communication round. Then, training the neural network and dynamic pruning components together using the local dataset. After training is finished, the client will upload the neural network's parameter to the server, and the dynamic pruning component's parameter $w_g$ is stored locally. Therefore, updating models from the clients to the server would not incur extra communication costs.
The loss function is defined in equation~\ref{eq:loss}.

\begin{equation}\label{eq:loss}
    \mathcal{L}(w,w_g;\mathcal{B}) = \mathcal{L}_{nn} + \mathcal{L}_g
\end{equation}
where the $\mathcal{L}_{nn}$ is the Cross-Entropy loss of the neural network, the $\mathcal{L}_g$ is the gate loss defines in section~\ref{sec:dpg}, and the $\mathcal{B}$ is the local data.

\section{Experiment}

\subsection{Implementation details}
\textbf{Dataset.}\
The experiments are done on CIFAR-10~\cite{krizhevsky2009cifar}. We do not use LEAF~\cite{caldas2019leaf} benchmark because they contain tiny datasets (such as EMNIST~\cite{cohen2017emnist}) that are trivial to modern CNNs. We used distribution-based label imbalance to simulate label imbalance between each client~\cite{li2021federated}. Here, each client is allocated a proportion of the samples of each label according to Dirichlet distribution (with concentration $\beta$). Specifically, we sample $p_k \sim Dir_N(\beta)$ and allocate a $p_{k,j}$ proportion of the instances of class $k$ to party $j$. Here we choose the $\beta = 0.1$.
Moreover, to test the \proj's dataset-aware pruning policy on each device, we further allocate a local validation dataset for each of them.

\textbf{Model architectures.}\ Two large modern CNN architectures are evaluated: VGG-11~\cite{simonyan2015vgg} and ResNet-32~\cite{he2016resnet}. Simple models (like CNN with several convolutional layers in LEAF) are not chosen because the effect of pruning cannot be realized in those cases.

\textbf{Federated learning settings.}
We experiment with different numbers of clients from 10, 30, 50 to 100 with sample rates 100\%, 40\%, 70\%, and 40\% respectively. For each training round, there are 10 epochs of local updates and in each epoch, the cosine learning rate decay is applied. We use the SGD optimizer~(mini-batch gradient descend), where the $\alpha=5~\times 10^{-3}$, the learning rate is $0.1$, batch size is 64, and the weight decay is $5~\times 10^{-4}$.

\subsection{Comparison with SoTA}

The goal of this part is to assess \proj's learning efficiency by looking at the relationship between communication rounds and the model's Top-1 accuracy, i.e., the answer that has the highest possibility of being correct must be the highest probability forecast. We train the VGG-11~\cite{simonyan2015vgg} on CIFAR-10~\cite{krizhevsky2009cifar} until convergence on the global model and compare the results with SoTA~(i.e.,  FedNova~\cite{wang2020fednova}, FedAvg~\cite{mcmahan2017fedavg}, FedProx~\cite{li2020fedprox}, and SCAFFOLD~\cite{karimireddy2020scaffold}).

\begin{figure*}
\begin{center}
 \vspace{-6 px}

\centerline{\includegraphics[width=.7\linewidth]{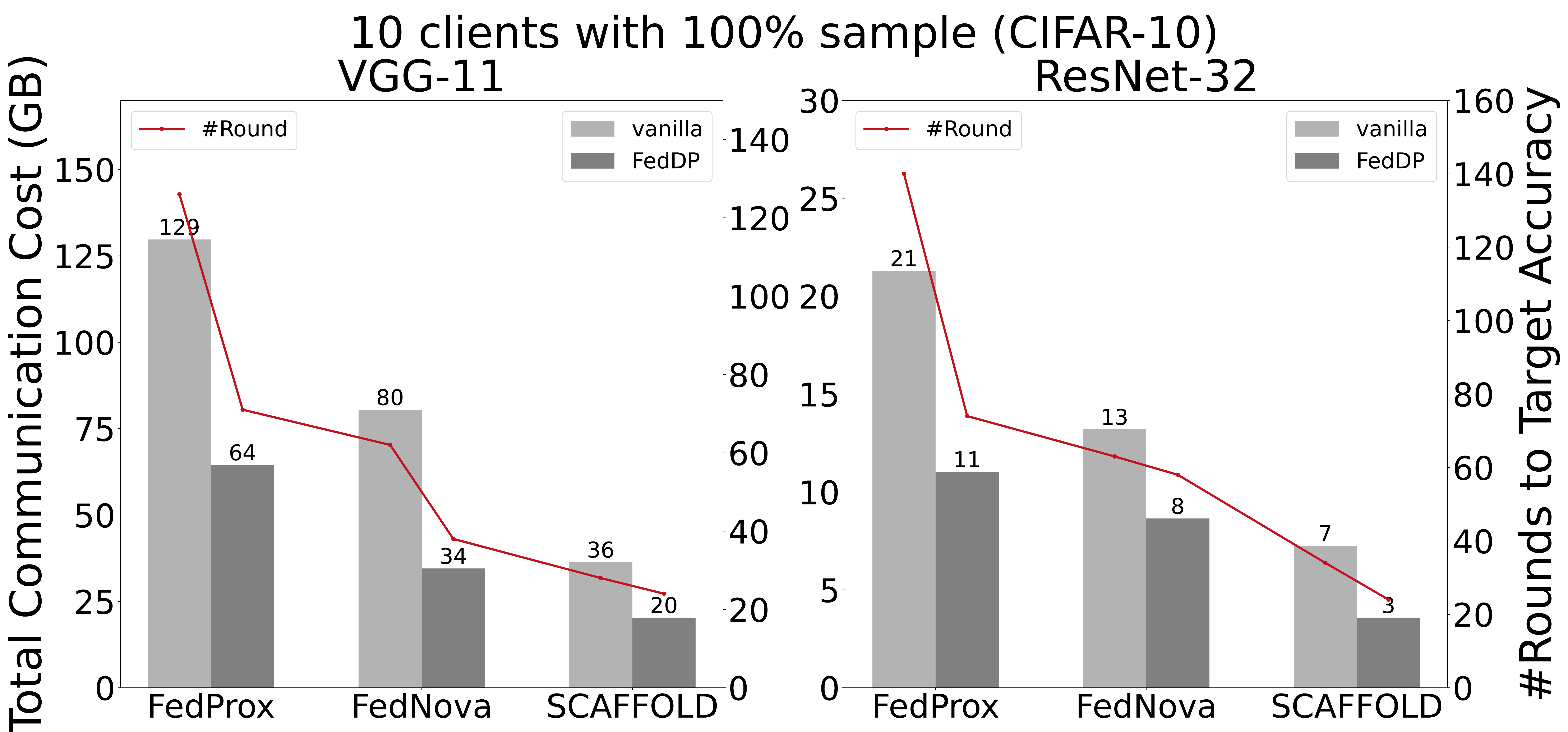}}

\caption{Communication cost to achieve target $78\%$ accuracy.}
\vspace{-6 px}
\label{fig:comm_cost}
\end{center}
 \end{figure*}

Figure~\ref{fig:scaffold_dp} shows the results of \proj on different configurations. More specifically, with 10 clients, \proj can yield a comparable training process with other methods. That means deploying dynamic pruning components in the local clients does not add extra challenges to the federated learning process. 
For the over-parameterized neural networks, such as VGG-11, \proj's highest accuracy is similar to that of SCAFFOLD (which has the best converging accuracy of all vanilla FL methods), where the accuracies achieved by \proj are 83.52\% and 84.21\% respectively. Moreover, the effectiveness of our proposed method is demonstrated in more complex FL settings, i.e., in which we scaled the total client numbers to 30, 50, and 100 with different sample ratios.

\begin{figure*}
 \begin{center}



\centerline{\includegraphics[width=\linewidth]{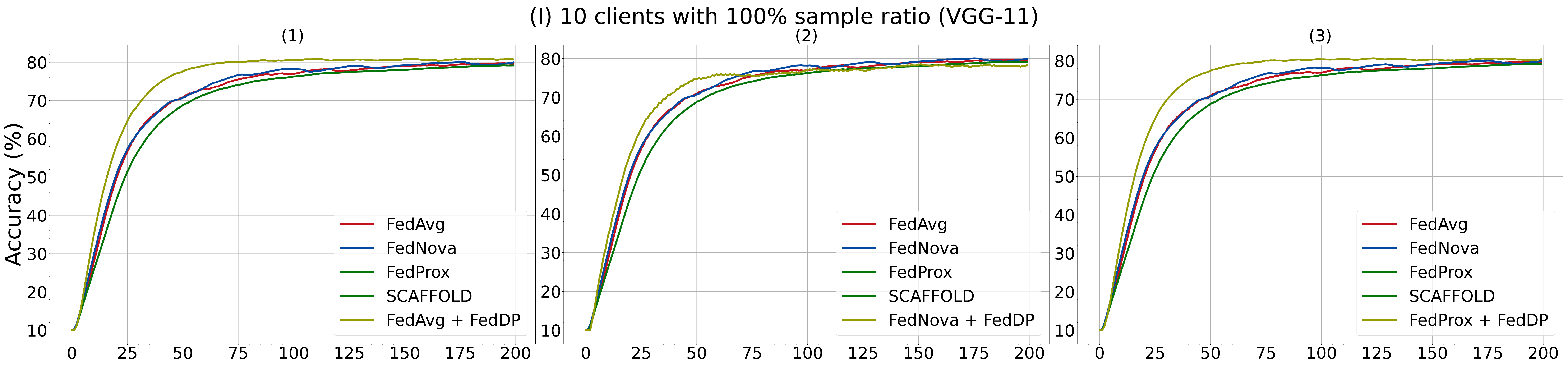}}
\centerline{\includegraphics[width=\linewidth]{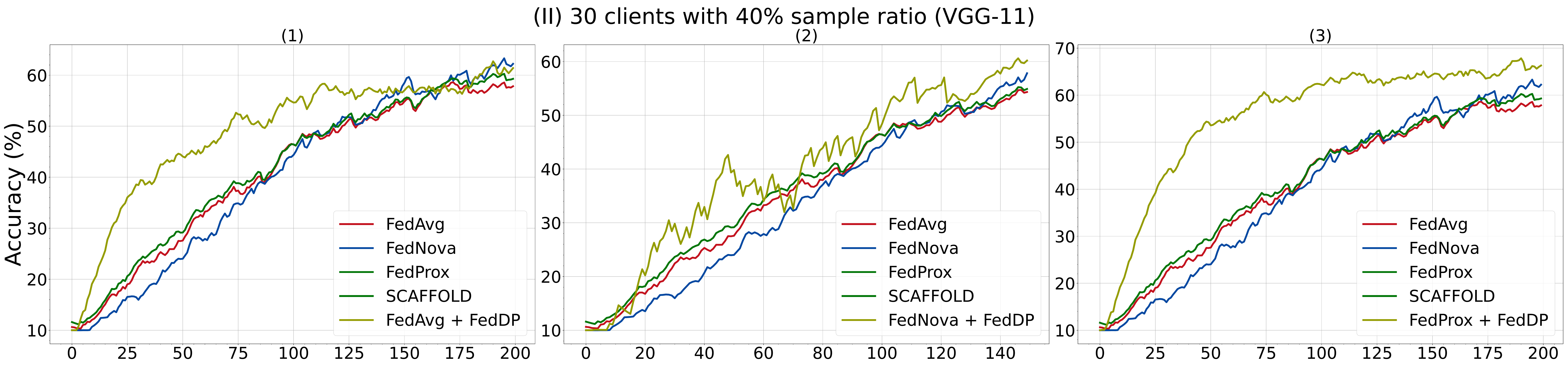}}
\centerline{\includegraphics[width=\linewidth]{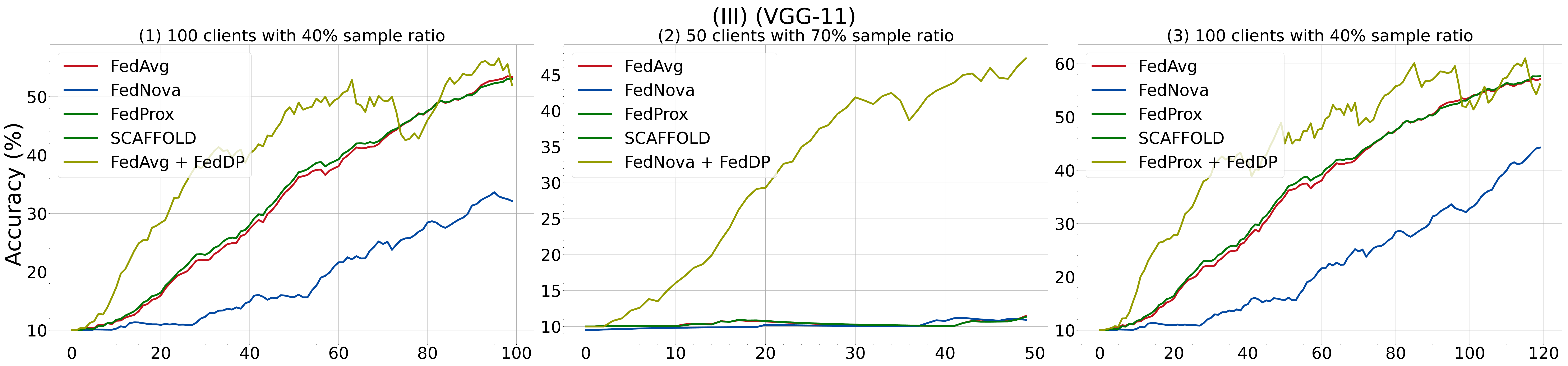}}

\caption{Varieties of FedDP with SoTA: the top-1 average test accuracy vs. communication rounds.}
 \vskip -0.3in
\label{fig:vgg_cifar}
\end{center}
 \end{figure*}

With the same dataset, increasing the number of clients will naturally cause more heterogeneity. The results indicate that \proj is suitable for dealing with non-IID data.
Experiment results in Figure~\ref{fig:vgg_cifar} show that with a more complex FL setting, the \proj consistently outperforms SoTAs, especially in the early stage.
In the case of 30 and 100 clients \footnote{There are some biased results shown in Fig.~\ref{fig:vgg_cifar}, where the SCAFFOLD\cite{karimireddy2020scaffold} diverges with gradient explosion in our experiment settings~\cite{li2022NonIIDBench} when there are more than 10 clients.}, 
for VGG-11, \proj outperforms all SoTA FL methods by a significant margin. Noticeably, \proj requires much fewer rounds in the beginning to reach the target accuracy and then saturates. The difference is more pronounced in 50 clients setting where \proj needs only 25 rounds to reach 35\% accuracy. 


We found it is because of the fact that over-parameterized neural networks are easily overfitting on local non-IID datasets by analyzing the training logs, which drifts local optima with respect to the global gradient optimum and thus makes the training process challenging. We claim that dynamic pruning is a good way to eliminate over-fitting.

\subsection{Inference acceleration}
We further evaluate the inference acceleration of \proj. 
One key contribution of \proj is to accelerate the inference time of deployed models. When doing inference in the local clients, we forward the input instance to the selected salient sub-model, accelerating the inference time and reducing computational consumption. Table~\ref{tab:flops} shows the inference acceleration status after training is finished. \proj notably reduced the FLOPs~(floating point operations per second) in all the evaluated models. For VGG-11 and ResNet-32, the FLOPs reduction among all clients is $50\%$ lower
than the original model, while the client models have a relatively low sparsity ratio~(the sparsity ratio represents the ratio of salient parameters compared to the entire model parameters).

The total FLOPs consumption of our best performing method~(SCAFFOLD) with and without dynamic pruning is recorded in Table~\ref{tab:flops}. The original VGG-11 has 15.48 billion FLOPs and under $2\times$ acceleration, we reduced 7.36 billion FLOPs~($50\%$ FLOPs compared to the original model). Similarly, the FLOPs of ResNet-32 have been reduced from 7.36 billion to 3.68 billion.

\begin{table}
  \caption{Comparison table for FLOPs.}
  \label{tab:flops}
  \centering
  \begin{tabular}{ccc}
    \toprule
    Model  & Original FLOPs & FLOPs After Pruning \\
    \midrule
    VGG-11    & 15.48 G  & 7.74 G ($50\% \downarrow$)   \\

    ResNet-32    & 7.36 G & 3.68 G ($50\% \downarrow$) \\
   
    \bottomrule
  \end{tabular}
\end{table}

\subsection{Training efficiency}
One key issue to the federated learning poor convergence is the deployed model's overfitting on local non-IID data. To investigate if dynamic pruning can significantly reduce the chance and contribute to FL's training efficiency, in this section, we use SoTA FL methods and \proj to train a deployed model to a target accuracy, and analysis their communication and computational cost. In practice, we train the deployed model to reach 78\% accuracy. As Figure~\ref{fig:comm_cost} demonstrates, \proj with less probability to over-fitting can achieve the target accuracy much faster. 

Figure~\ref{fig:comm_cost} shows the training details on VGG-11 with and without pruning on three FL algorithms: FedAvg, FedNova, and SCAFFOLD. We can see that \proj reduces communication rounds to reach the target accuracy~($78\%$) in all methods while having lower total communication costs.

The function for computing the cost is defined as: 
\begin{equation*}
\begin{aligned}
\text{Cost} = \text{Number of Parameters} \times \text{Rounds to reach target} \\ \text{accuracy} \times \text{Number of clients} \times \text{Sample rate}
\end{aligned}
\end{equation*}

The total communication cost when training the VGG-11~\cite{simonyan2015vgg} and ResNet-32~\cite{he2016resnet} is recorded in Figure~\ref{fig:comm_cost}. We have applied the \proj to various FL methods. The results show our approach can significantly reduce the communication cost of all the FL methods. Especially, on FedProx~\cite{li2020fedprox}, by applying the \proj with VGG-11, 65 GB communication cost had been reduced, which is $2 \times$ less cost than the original FedProx. Additionally, we found the \proj achieves the best performance on the SCAFFOLD~\cite{karimireddy2020scaffold}, where training the VGG-11 and ResNet-32 only takes 20 GB and 3 GB communication cost respectively, which is significantly superior to others.

\section{Ablation study}
In this section, we investigate the effect of dynamic pruning on federated learning algorithms. In essence, we equip dynamic pruning with the rest of FL learning algorithms: FedAvg, FedNova, and FedProx. The results are summarized in Figure~\ref{fig:vgg_cifar}.

Across all FL methods with different numbers of clients and sample rates, each method when paired with FedDP shows strong performance. Specifically, with 10 clients settings, FedDP converges at the same accuracy as others while needing fewer rounds. Moreover, dynamic pruning can markedly improve final accuracies for all methods in the cases of 30, 50, or 100 clients (meaning more heterogeneity in local datasets). The general trend throughout experiments is that FedDP helps to achieve a target accuracy faster in the beginning and it's converged accuracy is at least as good as the vanilla methods. This has demonstrated the applicability of FedDP over many SoTA federated learning methods.

\section{Conclusion}
In this paper, we study non-IID data as one key challenge in federated learning and develop pruning methods to customize each device's model further. We show that with dataset-aware dynamic pruning, the quality of the model is almost always better than the original across state-of-the-art federated learning approaches. Furthermore, while dynamic pruning can adapt different pruning policies for each client's data, our method reduces communication costs as well as FLOPs. This work is a starting point for addressing network pruning on the heterogeneity problem in federated learning. There are still many challenges left to make the idea more robust and applicable in real-life scenarios. 
In future work, we want to extend dataset-aware dynamic pruning towards communication effectiveness to improve the robustness of the idea with federated learning and enable additional algorithms for faster and better converging. More extensive experiments on different datasets and models will be another direction for our method.  
     


\medskip

\bibliographystyle{IEEEtran}
\bibliography{IEEEabrv,reference}

\end{document}